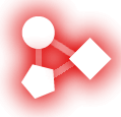

# Robotics Rights and Ethics Rules

**Tuncay Yigit**
*Suleyman Demirel University, Turkey*
*tuncayyigit@sdu.edu.tr*

**Utku Kose**
*Suleyman Demirel University, Turkey*
*utkukose@sdu.edu.tr*

**Nilgun Sengoz**
*Mehmet Akif Ersoy University, Turkey*
*nilgunsengoz@mehmetakif.edu.tr*

## Abstract

It is very important to adhere strictly to ethical and social influences when delivering most of our life to artificial intelligence systems. With industry 4.0, the internet of things, data analysis and automation have begun to be of great importance in our lives. With the Yapanese version of Industry 5.0, it has come to our attention that machine-human interaction and human intelligence are working in harmony with the cognitive computer. In this context, robots working on artificial intelligence algorithms co-ordinated with the development of technology have begun to enter our lives. But the consequences of the recent complaints of the Robots have been that important issues have arisen about how to be followed in terms of intellectual property and ethics. Although there are no laws regulating robots in our country at present, laws on robot ethics and rights abroad have entered into force. This means that it is important that we organize the necessary arrangements in the way that robots and artificial intelligence are so important in the new world order. In this study, it was aimed to examine the existing rules of machine and robot ethics and to set an example for the arrangements to be made in our country, and various discussions were given in this context.

## 1. INTRODUCTION

'Artificial Intelligence', wrapping up today's world, has begun to progress by putting its own rules in a long and painful development process. ((Rashevsky, 1943),(Hebb, 1949), (Krizhevsky, Sutskever & Hinton, 2017)). Taken by John McCarthy, Marvin Minsky, Nathaniel Rochester, and Claude Shannon was accepted as the name of a new research discipline in the 1956 Dartmouth summer research project in a two-month study on artificial intelligence. (McCarthy et al., 1956). Even though McCarthy and his colleagues had hoped for great progress toward human-level machines at that time, this hope could not be met. The ability to make robots equivalent to the human brain has been found in all areas of life, even though it was only a dream for that period, until scientific research on artificial intelligence, philosophical writings and even science fiction films.

The idea that most new innovations in technology can substitute for people on the field has already begun to be applied among researchers. The robots, which are used to minimize the human error in production processes in factories, have become a new source of anxiety due to the idea that they may lead to unemployment in both developed and developing economies. The cost of technological evolution and robots in the robots has decreased compared to the previous years and they have come out of factories and spread to most areas of our lives the first robot with humanoid design was Honda ASIMO (Hirai, Hirose, Haikawa & Takenaka, 1998)followed by NAO (Gouaillier et al., 2009), LOLA (Ulbrich, Buschmann & Lohmeier, 2006) and WABIAN-2 (Yu Ogura et al., 2006). The ever-evolving robot industry has advanced to the ATLAS (Kuindersma et al., 2015) robot, which can finally flipped in support of artificial intelligence.

The rapid evolution of both the algorithmic level and the computer hardware, while trying to make people's lives easier, reveals some moral dilemmas on the other hand. Tesla's CEO, Elon Musk, the world's leading producer of electric vehicles and energy storage products, said Artificial Intelligence (AI) is a threat to human civilization. He also pointed out that artificial intelligence is a security problem because of its power density and it is such a great power that it is not right that only a few people working at Google can control it without the supervision and control mechanism ("AI will probably destroy humans, Elon Musk warns", 2018). The following sections of this article will focus on this topic in detail. But if we return to the day, a video was uploaded to YouTube that is a social networking site, by Boston Dynamics which is a Robot-producing company, in 2015. In this video, one of the employees kicked a dog to show how he could protect the dog-like robot 'Spot' balance. This has upset most people and has characterized this as cruelty and immorality. In another video, the same company employee pushed him through a long carton tube to show the ability of a robot named 'Atlas' to retreat. These videos, which are often shared on social media, have been subjected to the same reaction. In an article written by CNN writer Phoebe Parke, it turns out that as the days of robots began to look more and more similar to people, any violence against them caused them to become more disturbed and to react. In the same article Noel Sharkey, Professor of Artificial Intelligence and Robotics at the University of Sheffield in England, added the following words: *"The only way it's unethical is if the robot could feel pain."* (Phoebe Parke, 2018).

The rise of artificial intelligence (AI) brings not only technical challenges, but also significant legal and ethical challenges for the machine-related society, especially autonomous weapons and self-sustaining automobiles. The mission of both morality and law may be to discover how best to make life in a society and to see that society develops when we know certain truths. Laws that changed in accordance with the time first focused on human rights and continued with animal rights. As the effectiveness of AI and Robotic systems gradually spread to all areas of our lives, a number of ethical and legal rules have started to be defined in some measure.

Because of the artificial intelligence and the robots spreading with my life so intensely with the developing technology, some ethical problems have emerged in the society. The content of the article follows that remains in the context of the purposes of this study is organized as follows. The next chapter deals with the concept of ethics, in the next section, researches on Artificial Intelligence and Ethics are given in detail. In the final section, the conclusion part of the study is included.

## 2. CONCEPT OF ETHICS

Ethics is a thought process rather than a prescriptive process. In any case, one decides what is good or bad after a moral, legal, and cultural filter. In other words, ethics is a relative concept and activates a process mechanism based on the ecosystem it is in. And the key word in this mechanism is the consciousness. Having an ethical value is essentially a 'consciousness'. Every individual tries to make decisions that we call 'right' or 'wrong', taking into account the ethical rules of which he/she is. But we must never forget that there is not a single 'right' or a single 'wrong'. Ethical values only determine the boundaries of the circle, the principles applicable to the situation are dependent on one's own initiative.

There are three main fields of study in today's well-known ethics field;
- *Meta-ethics*, concerns the theoretical meaning of moral propositions, their references, and how their true values (if any) can be determined. It is ethical, focusing on axiomatic concepts used by normative ethics and applied ethics. "What is wrong" and "right" means what it means and how to apply them.
- *Normative ethics,* relate to practical means of defining a moral action line, that is, aiming to judge a person or action with certain moral theories.
- *Applied ethics,* focuses on real life situations related to what a person has to do (or is allowed to do) in a particular situation or in a specific field of action (Lin, Abney & Bekey, 2014).

Artificial intelligence and the ethical values required to be given to robots direct us to the concept of applied ethics. This study will focus on this concept and try to propose a solution to the problems we may encounter in the real world. Classification using artificial intelligence algorithms and data available in video / video processing, decision making, etc. how can we encode the essence of human beings to distinguish the most basic and machine from the ethical concept into an artificial intelligence algorithm? How can ethical concepts be developed that have the same qualities that everyone can accept on a global scale, even when ethical rules can be different even within the same country? If we consider cultural, moral and legal situations, what will we do with it and in what language? One of the most important problems is how can we protect the security if such coding can take place? With robots becoming active in our life, such and such problems have surfaced on the surface of water, and researchers and academics have begun to focus on ethical issues. In this context, it was thought that it could cause not only the researchers of computer science but the subject to be exposed to a vicious circle and they were involved in this common problem in social and human sciences researchers. (Figure 1).

The first publication, directly determining the basis of the robotics ethic, was a science fiction and short story 'Runaround' (story) written by Isaac Asimov in 1942, in which the three famous Robotics Rule took place. These three laws were constantly changed by Asimov in the context of science fiction work, and a fourth or zero law was added before the first three. The first rule is that a robot can not harm a person or cause a person to be damaged by being immobile. According to the second rule, a robot must obey the orders given by humans, so long as these orders do not contradict the first rule. The third rule relates to the fact that a robot must protect its own existence as long as it does not contradict the first and second rule (Campbell, 1964). Asimov, in his book Robots and Empire, published in 1985, has a numbered rule of zero. Accordingly, a robot can not harm humanity and can not allow humanity to be damaged by its inaction (Pagallo, 2013).

*(Left blank for the Figure 1. Figure 1 is on the next page.)*

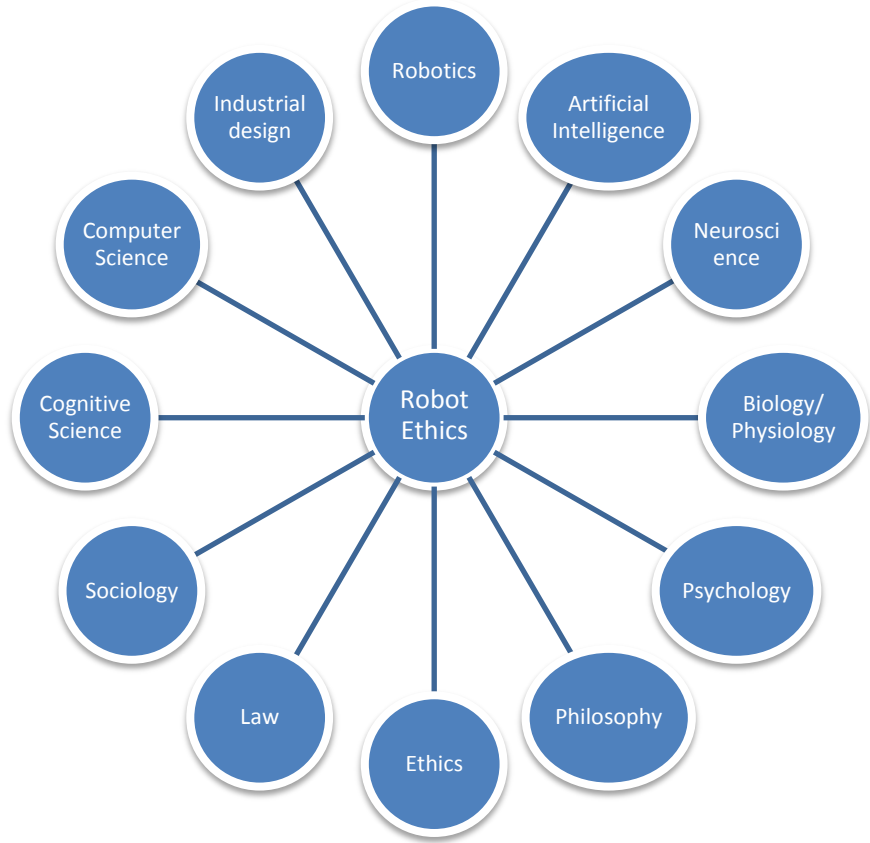

**Figure 1.** Disciplines on Robotics (Veruggio, 2006)

## 3. ARTIFICIAL INTELLIGENCE AND ETHICS

In January 2004, the Scuola di Robotica - Scuola Superiore Sant'Anna, the Pisa Art Institute and the Roman Pontificia Accademia della Santa Croce organized the Roboethics First International Symposium in cooperation with the Institute of Roman Theology. Together with the robot scientists, philosophers, jurists, sociologists, anthropologists and moralists are invited to contribute to the establishment of ethics bases in the design, development and use of robots. The anthropologist Daniela Cerqui set out three main ethical positions, two days of intense debate;

1- Those who are not interested in ethics; do not think that their actions are strictly technical and that they are a social or moral responsibility in their work.
2- Those interested in short-term ethical questions; According to this profile, the questions are expressed as "good" or "bad" and refer to some cultural values. For example, they think that robots must stick to social contracts. This includes 'showing respect' and helping people in various areas such as enforcing laws or helping seniors. Such thoughts are important, but reminiscent of the relative value of the values used to describe "bad" and "good" according to the contemporary values of industrialized countries.

Those who think in terms of long-term ethical questions, such as 'Digital division', ie South-North, or young-old; are aware of the gap between industrialized and poor countries and are wondering if the first material should change the ways of robotics development to be more beneficial for the South. The problem can not be formulated explicitly, so it can be considered as being implicitly covered (Gianmarco, 2005).

### 3.1. RoboEthics

The 'Roboethics' word was officially used for the first time, and the researchers began to move towards this area. In the same year (2004), a technical committee on 'Roboethics' was established by IEEE-RAS and also published the Fukuoka World Robot Declaration in Fukuoka, Japan. According to this declaration;

      I.   Expectations for the next generation of robots
  1. The next generation of robots will be 'common assets' living together with people
  2. The next generation of robots will help people both physically and psychologically
  3. The next generation of robots will contribute to both safety and peaceful society
      II.  Technology for creating new markets with new generation robots

  1. Solution of technical problems with efficient use of Special Zones for Robot Development and Testing
  2. Increasing the public acceptability of robots through the creation of standards and the improvement of the environment
  3. Promotion of adoption by promoting robots by public institutions
  4. Spread of new technologies related to robots
  5. Encouraging the development of robot technology by small businesses and encouraging entry into the robot business and providing active support for this effort by University-Industry-Government collaborations (Xu, Qian and Wu, 2004).

Following these developments, various conferences, workshops, etc. have been held on this subject for the following years. events were held. To put it briefly, the ICRA05 International Robotics and Automation Conference was held in Barcelona in 2005, focusing on the human-robot interaction and discusses how robots can be used for robotics and automation in the context of human approach, human-robot interaction, at home, at work, in education and in other emerging fields. a workshop on the problems in their progress and experiences has been carried out (Anon, 2005).

### 3.1.1. RoboEthics workshops

The Euron Roboethics Atelier workshop in Genoa, Italy, together with the E.C. EURON (EUropean RObotics research Network) project co-ordinated by the Robotics School, has set up a network of work where many robotists come together to make better robots. However, this project not only carried out robot research and development efforts, but also carried out some studies that produced the first road map of the robotics and examined the legal processes of the work (Robotica, 2018). Again in Italy, the International Biomedical Robotics and Biosepathronic Conference (BioRob, 2007) was held in the city of Pisa and a symposium on 'Roboethics' was held about robotic researchers, biomedical engineers, neuroscientists, biologists and scientists discussing future research lines (Anon, 2006).

In the workshop presentations made at the Workshop "Ethics of Human Interaction with Robotic, Bionic and Artificial Systems: Concepts and Politics" supported by the ETHICBOTS Europe Project, ethical issues, which are covered by 50 participants, both technical and human sciences, which are grouped under 6 headings;
- Fair access to adaptive and intelligent machine resources;
- Machine autonomy and prudential policies;
- Responsibilities for human-machine negotiation and action;
- The protection and support of fundamental human rights;
- Individual and social influences of human-machine cognitive and emotional ties and
- Intercultural aspects and use of robot and softbot design (Ethics of Human Interaction with Robotic, Bionic, and AI Systems: Concepts and Policies, 2006).

This interdisciplinary work has once again revealed the importance of robotics. This was discussed in detail at the International Conference on Artificial Intelligence and Law at Stanford University (Anonn, 2007).

The International Conference on Computers and Philosophy in Europe has been explored in all aspects of "computational transformation" that emerges from the interaction of philosophy and computer discipline with researchers from philosophy, computer science, social sciences and related disciplines (Anon, 2007). In the following years, the term 'Roboethic' has become very popular and events have been organized (Computer Ethics: Philosophical Enquiry-CEPE, 2007; Association, 2008; Communications of the ACM, 2009; UKRE Workshop on Robot Ethics, 2013).

On October 25, 2017 in Riyadh, at the "Future Investment Summit", a major breakthrough took place on the subject of robotics and ethics. Saudi Arabia 'Sophia', a woman named Sofia, granted a civil rights to a robot (Dünya'nın Ilk Robot Vatandaşı Suudi Arabistanlı, 2017). This is because it is not clear whether or not Sofia means that it can vote or marry, or that it is not clear whether the closure of a deliberate system can be regarded as 'murder'; It was also controversial when the Saudi Government's lack of rights to women was taken into consideration (Stone, 2017).

## 3.1. Ethical Rules for Engineers

The concepts of professionalism and ethics are two inseparable parts. In this context, the basic principles adopted by the National Society of Professional Engineers (NSPE) in 2017 guide engineers on this subject (Code of Ethics, 2017).

**Table 1.** Accepted Ethical Rules for Engineers [30]

| Basic Principles |
| --- |
| To provide public safety, health and prosperity |
| Perform services only according to their jurisdiction |
| Publish public expressions only objectively and accurately |
| Acting faithfully for every employer or client |
| Avoid deceptive actions |
| To be honorable, responsible, ethical and legal to develop the dignity, respectability and benefit of the profession |

These constantly updated ethical rules are an important guide to engineers. Especially when designing artificial intelligence algorithms in the concept of robot ethics, human interaction robots should be more careful, society should always pay attention to its benefits and peace. Likewise, the way engineers interact with their surroundings should never forget that robots will shape their interaction with their surroundings.

## 4. CONCLUSION

How do we overcome the ethical issue of artificial intelligence-based robots that enter every field of our lives, from health, finance, law, to safety? Is it possible to model artificial intelligence-based machines that can react to moral subjects like a human being? Is it even possible to make robots that decide according to the cultural and moral system in which they are doing? How can being 'conscious' be taught to the machine? Complex questions and even more are one of the issues that human beings must solve in the coming years. As a result, humans should not forget that if the ethical rules make the algorithmic base computable at the bottom of the machine understandable, perhaps it will also have its own end.

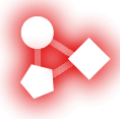